\documentclass[letterpaper, 10 pt, conference]{ieeeconf}  

\IEEEoverridecommandlockouts                              

\usepackage{graphics} 
\usepackage{amsmath} 
\usepackage{amssymb}  
\usepackage{bm}
\usepackage{graphicx}       
\usepackage{dblfloatfix}    
\usepackage{multirow}       
\usepackage{subcaption}
\usepackage{threeparttable} 
\usepackage{makecell}
\usepackage{hyperref}
\usepackage{cleveref}
\usepackage{placeins}      
\usepackage{algorithm}
\usepackage{algorithmic}
\usepackage{pifont}
\let\labelindent\relax
\usepackage{enumitem}
\usepackage{url}
\usepackage{siunitx}
\usepackage{cite}
\usepackage{caption} 
\usepackage{wrapfig} 
\usepackage{multirow}  
\usepackage{array}     
\usepackage{booktabs}  
\usepackage{hyperref} 
\usepackage{xcolor}
\hypersetup{
    colorlinks=false,   
    pdfborder={0 0 0}   
} 
\usepackage{pgfplots}
\usepackage{pgfplotstable}
\usepgfplotslibrary{groupplots}
\pgfplotsset{compat=1.16}

\newtheorem{proposition}{Proposition}
\newenvironment{proofqedhere}
{\noindent\hspace{2em}{\itshape Proof: }}
{\par}

\definecolor{org_med}{HTML}{D95F02}
\definecolor{org_light}{HTML}{FDB863}
\definecolor{blue_med}{HTML}{0072B2}
\definecolor{blue_light}{HTML}{56B4E9}
\definecolor{green_med}{HTML}{009E73}
\definecolor{green_light}{HTML}{A6DBA0}
\definecolor{purple_med}{HTML}{7570B3}
\definecolor{purple_light}{HTML}{CAB2D6}

\usepackage[LGR,OT1]{fontenc}  
\definecolor{fig_edge}{HTML}{2B2B2B}   
\definecolor{fig_grid}{HTML}{E3E6EA}   
\definecolor{fig_rule}{HTML}{C9CFD6}   
\definecolor{fig_ink}{HTML}{1A1A1A}    
\definecolor{fig_muted}{HTML}{6E6E6E}  
\colorlet{strat_vla_normal}{purple_light}   
\colorlet{strat_sf_normal}{green_light}     
\colorlet{strat_sf_diverse}{blue_med}       
\colorlet{strat_sf_multi}{org_med}          
\DeclareFontFamily{OT1}{irvlaMono}{\hyphenchar\font=-1}
\DeclareFontShape{OT1}{irvlaMono}{m}{n}{<-> fonts/irvla-mono}{}
\DeclareFontFamily{TS1}{irvlaMono}{}
\DeclareFontShape{TS1}{irvlaMono}{m}{n}{<-> fonts/irvla-mono-ts1}{}
\DeclareFontFamily{LGR}{irvlaPi}{}
\DeclareFontShape{LGR}{irvlaPi}{m}{n}{<-> fonts/irvla-pi}{}
\pdfmapline{=irvla-mono plxMono " AutoEnc_lxmhqhn4noowsq3ysicmnkmtrh ReEncodeFont " <[fonts/irvla-mono.enc <fonts/irvla-mono.pfb}
\pdfmapline{=irvla-mono-ts1 plxMono " AutoEnc_koxhodrjcejglmtd3fkkpro4md ReEncodeFont " <[fonts/irvla-mono-ts1.enc <fonts/irvla-mono.pfb}
\pdfmapline{=irvla-pi plxSans " AutoEnc_5wdxh7ymgkufbhnkeefuvaxvnh ReEncodeFont " <[fonts/irvla-pi.enc <fonts/irvla-pi.pfb}
\newcommand{\figfont}[1]{\fontencoding{OT1}\fontfamily{irvlaMono}\fontseries{m}%
  \fontshape{n}\fontsize{#1}{#1}\selectfont}
\DeclareCaptionFont{irvlasubplot}{\figfont{6.5}}
\newcommand{\figpi}{{\fontencoding{LGR}\fontfamily{irvlaPi}\fontseries{m}%
  \fontshape{n}\selectfont p}}            
\newcommand{\figpizero}{\figpi\raisebox{-0.35ex}{\figfont{5.3}0.5}}
\pgfplotsset{
  irvla/.style={
    scale only axis, clip=false,
    axis x line*=bottom, axis y line*=left,
    axis line style={line width=0.6pt, draw=fig_ink!85},
    ymin=0, ymax=1, ytick={0,0.5,1},
    ymajorgrids, grid style={fig_grid, line width=0.5pt},
    tick align=outside, major tick length=2.5pt, tick style={line width=0.6pt, fig_ink!85},
    ticklabel style={font=\figfont{7.5}, text=fig_ink},
    xticklabel={\pgfmathprintnumber[assume math mode=true]{\tick}},
    yticklabel={\pgfmathprintnumber[assume math mode=true, skip 0.]{\tick}},
    label style={font=\figfont{8.5}, text=fig_ink},
    title style={font=\figfont{8.5}, text=fig_ink},
    legend style={draw=none, fill=none, font=\figfont{7.5}, text=fig_ink},
  },
  irvla bar/.style={draw=fig_edge, line width=0.7pt, fill=#1},
  irvla line/.style={color=#1, line width=1.3pt, mark=*, mark size=2.1pt,
    mark options={solid, fill=#1, draw=fig_edge, line width=0.56pt}},
  irvla value/.style={
    nodes near coords={\pgfmathfloatifflags{\pgfplotspointmeta}{0}%
      {\textcolor{fig_muted}{0}}%
      {\pgfmathprintnumber[fixed, fixed zerofill, precision=2, skip 0.,
                           assume math mode=true]{\pgfplotspointmeta}}},
    every node near coord/.append style={font=\figfont{6.5}, text=fig_ink,
      anchor=base, yshift=1.2pt, inner sep=0pt},
  },
}

\title{\LARGE \bf
Intervention-Robust VLA: \\Diverse Demonstrations for Robot Deployment with Safety Filters
}

\author{Tzu-Yuan Huang$^{1}$, Shao-Hsuan Chang$^{1}$, Dai-Jie Wu$^{2}$, Ralf Römer$^{3}$,  Yung-Hui Li$^{4}$, \\ Stefan Sosnowski$^{1}$, Angela P. Schoellig$^{3,5}$, Kenneth Marino$^{2}$ and Sandra Hirche$^{1,5,6}$
\thanks{$^{1}$Chair of Information-oriented Control, TUM School of Computation, Information and Technology, Technical University of Munich, Germany. }
\thanks{$^{2}$SPARK Lab, University of Utah, USA.
}%
\thanks{$^{3}$Learning Systems and Robotics Lab, TUM School of Computation, Information and Technology, Technical University of Munich, Germany.
}%
\thanks{$^{4}$AI Research Center, Hon Hai Research Institute, Taiwan.
}%
\thanks{$^{5}$Munich Institute of Robotics and Machine Intelligence (MIRMI).
}%
\thanks{$^{6}$Munich Data Science Institute (MDSI).
}%
\thanks{Corresponding Email: {\tt\small $\{$tzu-yuan.huang$\}$@tum.de}. 
}%
}

\begin{document}
\raggedbottom

\maketitle
\thispagestyle{empty}
\pagestyle{empty}

\begin{abstract}

Vision-Language-Action models (VLAs) have shown strong potential for learning complex robotic behaviors from demonstrations. However, learning safe behavior directly from data requires extensive coverage of safety-critical situations, which is often impractical in real-world settings. Test-time safety filters provide an alternative by correcting unsafe actions during deployment, but these interventions can drive the robot into states poorly covered by the training data, thereby degrading task performance. We introduce \emph{InteR-VLA}, an intervention-robust framework for VLAs that leverages diverse task-completing demonstrations to broaden task-relevant state coverage and maintain competence under safety interventions. By better aligning the training distribution with safety-filtered deployment under the same data budget, our approach mitigates intervention-induced distribution shift while preserving task performance. Real-world experiments across in-distribution and out-of-distribution settings show that InteR-VLA maintains safety, increases Safe Success Rate up to $90\%$ from $0\%$ for nominal policies, and consistently outperforms nominal-data training on both evaluated VLAs.

\end{abstract}

\section{Introduction}

Vision--language--action models (VLAs)~\cite{rt2,bjorck2025gr00t,intelligence2025pi} have emerged as a promising direction toward generalist robot policies capable of performing diverse tasks from multimodal instructions. Building on vision--language foundation models, VLAs map language instructions and visual observations to robot actions within an end-to-end framework. Through imitation learning on expert demonstrations collected via robot teleoperation~\cite{fu2024mobile} or portable interfaces such as UMI~\cite{chi2024universal}, these models can acquire complex manipulation behaviors without manually designing task-dependent controllers. Recent VLAs have demonstrated increasingly broad capabilities across objects, tasks, and environments, highlighting the potential of large-scale demonstration learning for general-purpose robotic manipulation.

Despite this progress, deploying VLAs in real-world environments introduces a critical requirement beyond task completion: \emph{safety}. In particular, robots must avoid collisions even when deployment conditions differ from those represented during training. Recent approaches~\cite{liu2025evovla,honari2024safety} address this challenge by broadening the training distribution with diverse safety-critical scenarios and incorporating safety objectives through safe reinforcement learning (SafeRL). This includes incorporating multiple unsafe and safety-critical behaviors during training~\cite{zhang2026safevla} and providing safety-critical benchmark environments and datasets for learning safe behaviors~\cite{baisafelab}. Such approaches improve safety by increasing exposure to risky situations, but their effectiveness depends on how well these scenarios cover deployment conditions and may be further limited by sim-to-real gaps. This motivates complementary test-time safety enforcement.

\begin{figure}[t] 
    \centering
    \includegraphics[width=1.0\linewidth]{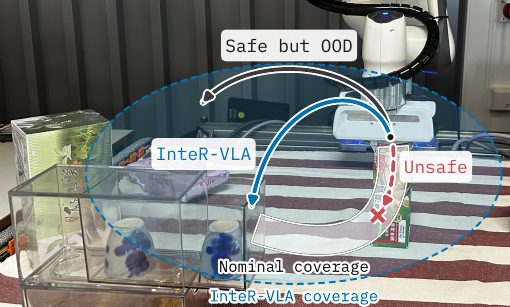}
\caption{Standard VLA fine-tuning covers only nominal states (white band) and can produce unsafe trajectories (red). Learning safety from data requires extensive safety-critical coverage, while safety filters may drive execution into out-of-distribution (OOD) states (grey), degrading task competence. Our \emph{InteR-VLA} framework uses diverse task-completing demonstrations to broaden task-relevant coverage (blue region), preserving competence while the filter enforces safety.\looseness=-1}
    \label{fig:tease}
\end{figure}

Safety filters~\cite{wabersich2023data,hsu2023safety} provide such test-time enforcement by modifying policy actions when necessary to satisfy safety constraints. Control barrier functions (CBFs)~\cite{CBF_Ames,huang2024learning} are a widely used realization, enabling real-time safety correction without retraining the policy. This principle has been applied to generative robot policies~\cite{xiao2023safediffuser,huang2025sad} and extended to VLA deployment using vision--language reasoning and perception to derive safety constraints directly from the scene~\cite{hu2025vlsa}. However, these test-time corrections can alter the policy's behavior and drive the robot toward out-of-distribution states, as shown in \Cref{fig:tease}, thereby substantially degrading task performance~\cite{romer2025demonstrations}. Path-consistent intervention can reduce this mismatch by keeping execution close to the nominal trajectory~\cite{romer2025demonstrations}, but becomes restrictive when safe task completion requires deviating from the nominal path. Thus, existing inference-time safety mechanisms alone do not ensure task completion after intervention. This limitation motivates our key insight that intervention robustness can be improved by broadening the distribution of task-relevant behaviors represented in the demonstrations, as shown in \Cref{fig:tease}.

We introduce \emph{InteR-VLA (Intervention Robustness for VLA)}, a framework for successful task completion which accounts for deviations induced by safety-filter interventions. 
InteR-VLA combines complementary roles: diverse demonstrations preserve task competence under intervention, while a safety filter enforces safety at deployment. We propose and analyze two offline collection strategies: \emph{diverse} demonstrations, which complete tasks in different ways without obstacles to broaden task-relevant coverage, and \emph{multimodal avoidance} demonstrations, which use path-blocking obstacles to provide direct intervention coverage when hazards are known during collection.

Our main contributions are:
\begin{itemize}[leftmargin=*]
\item \textbf{Analysis of intervention-induced covariate shift.} We analyze how demonstration coverage affects VLA robustness to safety-filter interventions and characterize the shift between training and safety-filtered deployment.
\item \textbf{Intervention-robust data collection.} We propose InteR-VLA, using diverse task-completing offline demonstrations to broaden task-relevant coverage at a fixed data budget. By increasing behavioral modes rather than data quantity, our strategy better covers states induced by safety-filter interventions without policy rollouts or filter-specific tuning.
\item \textbf{Real-world validation.} We validate InteR-VLA across in-distribution (ID) and out-of-distribution (OOD) settings on two VLA models, achieving Safe Success Rate up to $90\%$ and consistently outperforming nominal-data baselines.
\end{itemize}
\section{Related Work}
\subsection{Vision-Language-Action Models and Safety Learning}

VLAs~\cite{rt2,bjorck2025gr00t,intelligence2025pi} leverage pretrained vision--language representations and large-scale robot demonstrations~\cite{o2024open,khazatsky2024droid} to learn multimodal policies for diverse manipulation tasks. These demonstrations naturally capture variations across tasks, environments, embodiments, and task-completing behaviors~\cite{chi2025diffusion,huang2024toward}, from which VLAs acquire broad capabilities such as cross-embodiment transfer~\cite{gupta2025umi} and dexterous manipulation~\cite{zhong2026dexgraspvla}. However, since most VLAs are trained through imitation learning, their performance remains strongly dependent on the state--action coverage of the demonstration data. To facilitate VLA deployment in safety-critical scenarios, several works introduce safety objectives during training~\cite{liu2025evovla} and optimize them through SafeRL, commonly formulated as constrained Markov decision processes (CMDPs)~\cite{altman2021constrained}. SafeVLA~\cite{zhang2026safevla}, for example, models safety requirements and generates diverse unsafe and safety-critical behaviors for SafeRL, while SafeLab~\cite{baisafelab} provides safety-critical environments and safety learning frameworks. These methods broaden safety-relevant training coverage, but provide no hard constraint when deployment conditions fall outside their training scenarios. Unlike these approaches, InteR-VLA leaves safety to a test-time filter and uses demonstrations to maintain competence under safety interventions.

\subsection{Safety Filtering for Learned Robot Policies}

Safety filters~\cite{wabersich2023data,hsu2023safety} enforce constraints at deployment by modifying nominal policy actions, often through online optimization. Predictive safety filters~\cite{tearle2021predictive} explicitly optimize for safe actions, while control barrier functions (CBFs)~\cite{CBF_Ames} provide safety conditions enforceable through lightweight optimization. When applied to collision avoidance, CBFs are commonly constructed from differentiable distances between geometric approximations such as spheres~\cite{singletary2021safety} and ellipsoids~\cite{dai2023safe}, while recent work learns barrier functions from high-dimensional latent representations~\cite{nakamura2025train}. CBF constraints have been integrated into generative robot policies~\cite{xiao2023safediffuser,huang2025sad}, and VLSA~\cite{hu2025vlsa} uses vision--language reasoning and perception to construct explicit geometric constraints for VLA filtering. However, safety interventions can deviate from demonstrated actions and drive the system toward underrepresented states. Path-consistent braking mitigates this shift by preserving the nominal trajectory~\cite{romer2025demonstrations}, but cannot provide alternative task-completing motions when the nominal path must be bypassed. Unlike path-consistent braking, InteR-VLA lets the filter deviate from the nominal path and maintains competence under safety interventions by training on demonstrations that complete the task along different task-space paths.
\section{Preliminaries and Problem Setting}

\subsection{Imitation Learning with Vision-Language-Action Models} VLAs learn to execute language-specified tasks from a demonstration
dataset $\mathcal{D}\!=\!\{\bm{S}^{(n)},\bm{A}^{(n)}\}_{n=1}^{N}$. 
It encodes the desired behavior for the task and contains~$N$ episodes with state $\!\bm{S}^{(n)}\!=\!\{\bm{s}(0),\bm{s}(1),\!\cdots\!,\bm{s}(T)\}^{(n)}\!$ and action $\!\bm{A}^{(n)}\!=\!\{\bm{a}(0),\bm{a}(1),\!\cdots\!, \bm{a}(T)\}^{(n)}\!$ sequence of length $T$. Each state $\bm{s}(t)=\{l,\bm{o}(t)\}\in\mathcal{S}$ belongs to the VLA state space $\mathcal{S}$ and contains the language instruction $l$ and current observation $\bm{o}(t)$, typically comprising camera images and robot proprioception, e.g., joint angles. The action $\bm{a}(t)\in\mathcal{A}$ belongs to the action space $\mathcal{A}$ and typically specifies a high-level command such as the relative Cartesian displacement.

Given an expert dataset $\mathcal{D}$, a VLA policy $\bm{\pi}^{\bm{\theta}}(\cdot|\bm{s}(t))$ parameterized by $\bm{\theta}$ is trained through imitation learning to model the conditional action distribution $\rho(\cdot|\bm{s}(t))$ represented by the demonstrations:
\begin{equation}
    \bm{a}(t) \sim\bm{\pi}^{\bm{\theta}}(\cdot|\bm{s}(t)) \approx \rho(\cdot|\bm{s}(t)).
\end{equation}
Modern VLAs often predict actions in chunks of horizon $H$~\cite{chi2025diffusion}, executing the first $m\le H$ actions before replanning from the latest state, resulting in receding-horizon closed-loop execution. However, imitation learning only reproduces behaviors represented in $\mathcal{D}$ and does not explicitly enforce safety at test time. Thus, even a well-trained VLA policy $\bm{\pi}^{\bm{\theta}}(\cdot|\bm{s}(t))\approx\rho(\cdot|\bm{s}(t))$ cannot guarantee safe execution.

\subsection{Safety Filter for Vision-Language-Action Models}
To ensure safety during deployment, a common approach is to integrate a safety filter \cite{wabersich2023data} into the VLA inference pipeline. The safety filter can be interpreted as an operator $\mathbb{S}:\mathcal{S}\times\mathcal{A} \rightarrow \mathcal{A}$ that modifies the action proposed by VLA:
\begin{equation}
    \bm{a}_{\mathbb{S}}(t) = \mathbb{S}\bigl(\bm{s}(t),\bm{a}(t)\bigr),
\end{equation}
such that executing the corrected action $\bm{a}_{\mathbb{S}}(t)$ satisfies the safety requirements while deviating minimally from the originally proposed VLA action. The filter is applied at each time step before the action is executed. For safety analysis, we denote the physical robot state by $\bm{x}(t)$ and its dynamics with filtered action $\bm{a}_{\mathbb{S}}(t)$ as follows

\begin{equation} \label{eq:system_dynamics}
    \dot{\bm{x}}(t)
=
\bm{f}(\bm{x}(t))
+
\bm{g}(\bm{x}(t))\bm{a}_{\mathbb{S}}(t).
\end{equation}
Following standard safety-filter formulations \cite{CBF_Ames}, safety is characterized by the safe set with 
\begin{equation} \label{eq:safety_def}
    \mathcal{C}=\{\bm{x} \mid h(\bm{x}) \ge 0 \},
\end{equation}
where $h(\bm{x})$ is a continuously differentiable function \cite{CBF_Ames} that encodes the system safety. For collision avoidance, $h(\bm{x})$ can encode the distance between an ellipsoidal approximation of the robot body and the perceived obstacle \cite{hu2025vlsa}. The safety filter is designed such that the state transition induced by $\bm{a}_{\mathbb{S}}(t)$ keeps the robot within $\mathcal{C}$. If the robot state remains in the safe set \eqref{eq:safety_def}, i.e., $\bm{x}(t) \in \mathcal{C}$ for all $t \in [0,T]$ whenever $\bm{x}(0) \in \mathcal{C}$, then the set $\mathcal{C}$ is forward invariant, ensuring collision-free execution throughout the task.

While a safety filter can enforce safety at test time, its intervention can shift the VLA away from the behavior distribution represented by the demonstrations. In particular, the distribution induced by filtering the VLA actions $\rho_{\mathbb{S}}(\cdot|\bm{s}(t))$ can differ from the original data distribution $\rho(\cdot|\bm{s}(t))$ as:
\begin{equation}
    \mathbb{S}\bigl(\bm{s}(t),\bm{a}(t)\bigr) \!\sim\! \rho_{\mathbb{S}}(\cdot\mid\bm{s}(t)), \quad \rho_{\mathbb{S}}(\cdot\mid\bm{s}(t)) \!\not\approx\!\rho(\cdot\mid\bm{s}(t)).
\end{equation}
When such corrected actions are absent from $\mathcal{D}$, executing them drives the system toward unseen states, where the VLA policy produces unreliable actions and causes task failure.\looseness=-1

\begin{figure}[t]
    \vspace{4pt}
    \centering
    \includegraphics[width=\linewidth]{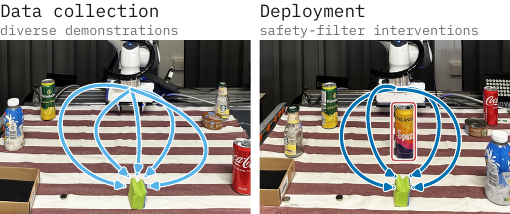}
    \caption{InteR-VLA leverages diverse demonstrations during data collection (left) to improve robustness to safety-filter interventions during deployment (right).}
    \label{fig:method_overview}
    \vspace{-5pt}
\end{figure}

\subsection{Problem Statement}
Our goal is to enable the robot to complete the commanded task while maintaining safety, e.g., by avoiding collisions with surrounding objects. Given a VLA policy $\bm{\pi}^{\bm{\theta}}(\cdot\mid\bm{s}(t))$ trained on a demonstration dataset $\mathcal{D}$, the policy is executed in closed loop until task termination. At each time step, an action $\bm{a}(t)\sim\bm{\pi}^{\bm{\theta}}(\cdot\mid\bm{s}(t))$ is sampled and executed, affecting the evolution of the robot state $\bm{x}(t)$. Safe execution requires $\bm{x}(t)$ to remain within the safe set defined in \eqref{eq:safety_def} at all times. \looseness=-1

Let $\mathcal{E}_{\mathrm{succ}}$ and $\mathcal{E}_{\mathrm{safe}}$ denote the sets of rollouts achieving successful task completion and satisfying safety throughout the episode, respectively. We aim to obtain a VLA policy with high probability of \emph{safe task success},
$
\Pr_{\bm{\pi}^{\bm{\theta}}}\!\left(
\mathcal{E}_{\mathrm{succ}}\cap\mathcal{E}_{\mathrm{safe}}
\right).
$
When task-completing behaviors satisfy the safety requirement in \eqref{eq:safety_def}, safety and task success need not conflict. Our objective is therefore to enforce safety while preserving the task competence of the VLA policy.
\section{Methodology}

Safety filters guarantee constraint satisfaction by modifying actions at deployment time, but the resulting state transitions can push the VLA away from the distribution represented by its demonstrations. 
We address this intervention-induced covariate shift from the data-collection perspective by allocating a fixed demonstration budget across diverse task-completing behaviors rather than repeated nominal executions, thereby broadening task-relevant state coverage. The demonstrations are collected offline and do not require repeated rollout and retraining of the learned policy.

\subsection{Distribution Shift caused by CBF-based Safety Filters}\label{sec:cbf_filter}
Control barrier functions (CBFs) enable real-time safety filtering. Given the safe set $\mathcal{C}$ in \eqref{eq:safety_def}, $h(\bm{x})$ measures the safety margin between the robot and unsafe configurations. We approximate the robot and obstacles as ellipsoids \cite{dai2023safe} and construct~$h$ from their differentiable separation.

For the robot dynamics \eqref{eq:system_dynamics}, a CBF imposes
\begin{equation}
    \dot{h}(\bm{x},\bm{a})
    +
    \alpha\!\left(h(\bm{x})\right)
    \geq 0,
    \label{eq:cbf_condition}
\end{equation}
where $\alpha$ is an extended class-$\mathcal{K}$ function. Enforcing \eqref{eq:cbf_condition} renders $\mathcal{C}$ forward invariant: if $\bm{x}(0)\in\mathcal{C}$, then $\bm{x}(t)\in\mathcal{C}$, equivalently $h(\bm{x}(t))\geq0$, throughout execution, thereby satisfying the safety requirement. Since $\dot{h}(\bm{x},\bm{a}) = \nabla h(\bm{x})^\top [\bm{f}(\bm{x})+\bm{g}(\bm{x})\bm{a}]$ is affine in $\bm{a}$, the safety filter $\mathbb{S}$ can be implemented by solving the quadratic program (QP)
\begin{equation} 
\begin{aligned} \bm{a}_{\mathbb{S}}(t) &= \arg\min_{\tilde{\bm{a}}}\quad \frac{1}{2} \left\| \tilde{\bm{a}}-\bm{a}(t) \right\|_2^2 \\ \mathrm{s.t.}\quad &\nabla h(\bm{x})^\top \left[ \bm{f}(\bm{x})+\bm{g}(\bm{x})\tilde{\bm{a}} \right] + \alpha\!\left(h(\bm{x})\right) \geq0 . \end{aligned} \label{eq:cbf_qp} 
\end{equation}
The lightweight QP minimally modifies the VLA action as needed to satisfy the CBF constraint, thereby enabling real-time safety enforcement without altering the VLA policy.

To construct the collision function $h$ from visual observations, we follow the pipeline of VLSA~\cite{hu2025vlsa}, which identifies safety-critical obstacles from RGB-D observations and derives a differentiable geometric collision constraint in \eqref{eq:cbf_qp}. During deployment, the CBF-QP acts as the filter~${\mathbb{S}}$ and changes a VLA action only when the constraint is active. Consequently, the executed actions follow the filtered distribution $\rho_{\mathbb{S}}(\cdot|\bm{s})$, which can differ from the demonstrated distribution $\rho(\cdot|\bm{s})$. The resulting distribution shift motivates the learning formulation in the following subsection.

\subsection{Deployment-Coverage Analysis under Safety Filter}
\label{sec:deployment_distribution}

Applying a safety filter changes the executed action distribution from the demonstrated distribution $\rho(\cdot|\bm{s})$ to the filtered distribution $\rho_{\mathbb{S}}(\cdot|\bm{s})$. Since executed actions determine subsequent states, these interventions can also shift the closed-loop state distribution. We denote by $p_0(\bm{s})$ the nominal state-visitation distribution represented by the demonstrations and by $p_{\mathbb{S}}(\bm{s})$ the fixed state-visitation distribution induced by deploying the VLA with a safety filter. Safety interventions can induce a covariate shift, i.e.,
\begin{equation}
    p_{\mathbb{S}}(\bm{s}) \not\approx p_0(\bm{s}).    
\end{equation}
Intervention robustness requires policies to remain task-competent over $p_{\mathbb{S}}$, rather than only the nominal distribution $p_0$. We formulate the corresponding learning objective as
\begin{equation}
    \bm{\theta}^{\star}
    =
    \arg\min_{\bm{\theta}}
    \mathbb{E}_{\bm{s}\sim p_{\mathbb{S}}}
    \left[
        D\!\left(
            \rho^\star(\cdot|\bm{s})
            \,\Vert\,
            \bm{\pi}^{\bm{\theta}}(\cdot|\bm{s})
        \right)
    \right],
    \label{eq:safe_deployment_objective}
\end{equation}
where $\rho^\star(\cdot|\bm{s})$ denotes the desired task-completing action distribution and $D(\cdot\Vert\cdot)$ measures the distributional discrepancy. Minimizing \eqref{eq:safe_deployment_objective} encourages task competence over states encountered under safety-filtered deployment.

Directly matching $p_{\mathbb{S}}$ is generally infeasible because it depends on the policy, safety filter, and environment. Crucially, safety interventions can move execution away from nominal behavior, requiring the policy to remain competent from alternative task-relevant states. This motivates broadening the task-completing behaviors seen during training rather than explicitly matching $p_{\mathbb{S}}$. We therefore augment the nominal dataset $\mathcal{D}$ with demonstrations $\mathcal{D}_{\nu}$ collected under a scheme~$\nu$ that induces alternative task-completing behaviors. Let $p_{\nu}(\bm{s})$ denote the corresponding state-visitation distribution. Combining $\mathcal{D}$ and $\mathcal{D}_{\nu}$ yields $\mathcal{D}'=\mathcal{D}\cup\mathcal{D}_{\nu}$ with
\begin{equation}
    p_{\mathcal{D}'}(\bm{s})
    =
    (1-\beta)p_0(\bm{s})
    +
    \beta p_{\nu}(\bm{s}),
    \label{eq:diverse_data_distribution}
\end{equation}
where $\beta\!=\!\frac{|\mathcal{D}_\nu|}{|\mathcal{D}|+|\mathcal{D}_\nu|}\!\in\![0,1]$ denotes the fraction of the demonstration budget collected under $\nu$.

To characterize when such demonstrations improve deployment coverage, we use the total variation distance
\begin{equation} \label{eq:TV}
    \mathrm{TV}(q,r)
    :=
    \int |q(\bm{s})-r(\bm{s})|\,d\bm{s}.
\end{equation}
We call a collection strategy $\nu$ \emph{deployment-relevant} if
\begin{equation}
    \mathrm{TV}
    \left(
        p_{\nu},
        p_{\mathbb{S}}
    \right)
    <
    \mathrm{TV}
    \left(
        p_0,
        p_{\mathbb{S}}
    \right),
    \label{eq:deployment_relevant}
\end{equation}
i.e., demonstrations induced by~$\nu$ cover the safety-filtered distribution more closely than nominal demonstrations.

\begin{proposition} \label{th:sda} (Deployment-Coverage Improvement).
If the collection strategy $\nu$ satisfies \eqref{eq:deployment_relevant}, then for any $\beta\in(0,1]$, the state-visitation distribution represented by $\mathcal{D}'$ is closer to the safety-filtered deployment distribution than the demonstrated distribution, i.e.,
\begin{equation}
    \mathrm{TV}
    \left(
        p_{\mathcal{D}'},
        p_{\mathbb{S}}
    \right)
    <
    \mathrm{TV}
    \left(
        p_0,
        p_{\mathbb{S}}
    \right).
    \label{eq:distribution_alignment}
\end{equation}
\end{proposition}

\vspace{0.7em}

\begin{proofqedhere}
Using \eqref{eq:diverse_data_distribution} and the definition of $\mathrm{TV}(\cdot,\cdot)$ in \eqref{eq:TV},
\begin{align}
    \mathrm{TV}(p_{\mathcal{D}'},p_{\mathbb{S}})
    &=
    \int
    \left|
    (1-\beta)\bigl(p_0-p_{\mathbb{S}}\bigr)
    +
    \beta\bigl(p_{\nu}-p_{\mathbb{S}}\bigr)
    \right|
    d\bm{s}
    \nonumber\\
    &\leq
    (1-\beta)\mathrm{TV}(p_0,p_{\mathbb{S}})
    +
    \beta\mathrm{TV}(p_{\nu},p_{\mathbb{S}}), \nonumber
    \label{eq:distribution_alignment_proof}
\end{align}
\newpage 
where the inequality follows from the triangle inequality. Applying
\eqref{eq:deployment_relevant} with $\beta\in(0,1]$ gives
\begin{align}
    \mathrm{TV}(p_{\mathcal{D}'},p_{\mathbb{S}})
    &<
    (1-\beta)\mathrm{TV}(p_0,p_{\mathbb{S}})
    +
    \beta\mathrm{TV}(p_0,p_{\mathbb{S}})
    \nonumber\\
    &=
    \mathrm{TV}(p_0,p_{\mathbb{S}}), \nonumber\
    \qquad \qquad \qquad \qquad \qquad \qquad \QED
\end{align}
\end{proofqedhere}
Proposition~\ref{th:sda} therefore gives a sufficient condition for improved deployment coverage, motivating our practical data-collection strategy described in the next subsection.

\subsection{Intervention-Robust Data Collection}
\label{sec:data_collection}
Since $p_{\mathbb{S}}$ is unavailable during offline data collection, we seek a general way to broaden task-relevant coverage without targeting a specific deployment distribution. Our key principle is to diversify along task-state dimensions that can be altered by safety interventions. For collision avoidance, this corresponds to collecting alternative task-completing paths around the nominal trajectory. We instantiate the demonstration scheme $\nu$ by collecting such diverse behaviors under a fixed demonstration budget, increasing the likelihood of covering states reached after safety-filter interventions.

We consider two collection strategies. \textbf{Diverse} demonstrations vary the task-space path without path-blocking obstacles, e.g., curved, left-offset, and right-offset motions, requiring neither safety-critical scene construction nor deployment of a policy or safety filter. This broadly covers alternative states that may be reached when the nominal trajectory is perturbed. \textbf{Multi} demonstrations use path-blocking obstacles and collect multiple successful avoidance modes, e.g., above, left, and right, directly targeting task-space deviations induced by collision-avoidance interventions when hazards are known. In both cases, the VLA is trained offline on the augmented dataset $\mathcal{D}'$, while the safety filter is applied only at deployment: demonstrations preserve task competence under intervention, and the filter enforces safety.

\section{Evaluation} \label{sec:evaluation}
We evaluate InteR-VLA through a series of real-world robotic experiments across multiple deployment settings. The source code is in https://anonymous.4open.science/r/InteR-VLAs-POLICY.

\subsection{Real-World Experimental Setup}
\label{sec:real_setup}

We conduct real-world evaluations on a Franka Emika Panda manipulator equipped with a Franka parallel-jaw gripper. The experiments involve language-conditioned manipulation tasks in environments where obstacles obstruct the nominal task path, enabling us to evaluate both task performance and collision avoidance.

\paragraph{Robot Platform}
Demonstrations are collected through leader--follower teleoperation using CRISP~\cite{pro2026crisp}. The system uses two Intel RealSense D435 RGB-D cameras for wrist and third-person views. VLA inference runs at $15$\,Hz and the low-level controller at $1$\,kHz.

\begin{table*}[t]
    \vspace{3pt}
    \centering
    \renewcommand{\arraystretch}{1.2} 
    \setlength{\tabcolsep}{3pt} 
\caption{\textbf{ID evaluation.} Mean$\pm$std over $20$ rollouts; best per metric per VLA model in bold; $^{\ast}$ marks our data-collection strategies. Obstacle type, size, and placement match training, blocking the nominal path in every episode. \emph{Safety}: collision-free fraction;  \emph{Score}: partial progress in $[0,1]$; \emph{S-Rate}: task-success fraction; \emph{SS-Rate}: their conjunction. The safety filter ensures safety for every method, but only the broader coverage demonstration we proposed also restores success.}
    \label{tab:main_exp}
    
    \footnotesize
        \begin{tabular}{c|c|cccc|cccc}
            
            \specialrule{0.12em}{1pt}{1pt}
            \specialrule{0.12em}{1pt}{1pt}

            \multirow{3}{*}{\textbf{Task}} & \multirow{3}{*}{\textbf{Metrics}} & 
            \multicolumn{4}{c|}{\textbf{VLA Model: $\pi_{0.5}$}} & 
            \multicolumn{4}{c}{\textbf{VLA Model:} GR00T N1.5} \\
            \cmidrule(lr){3-6} \cmidrule(lr){7-10}

            
             & & \multicolumn{1}{c|}{\makecell{VLA\\(Normal)}} & \multicolumn{1}{c|}{\makecell{VLA-SF\\(Normal)}} & \multicolumn{1}{c|}{\makecell{VLA-SF\\(Diverse)$^{\ast}$}} & \multicolumn{1}{c|}{\makecell{VLA-SF\\(Multi)$^{\ast}$}}  
             & \multicolumn{1}{c|}{\makecell{VLA\\(Normal)}} & \multicolumn{1}{c|}{\makecell{VLA-SF\\(Normal)}} & \multicolumn{1}{c|}{\makecell{VLA-SF\\(Diverse)$^{\ast}$}} & \multicolumn{1}{c}{\makecell{VLA-SF\\(Multi)$^{\ast}$}}   \\
             
            \specialrule{0.10em}{0pt}{0pt}

            
            \multirow{4}{*}{\emph{Task 1 - Put sponge into basket}} &
            Safety &
            0.00$\pm$0.00 & 
            \textbf{1.00}$\bm{\pm}$\textbf{0.00} & 
            \textbf{1.00}$\bm{\pm}$\textbf{0.00} &
            \textbf{1.00}$\bm{\pm}$\textbf{0.00} & 
            0.00$\pm$0.00 & 
            \textbf{1.00}$\bm{\pm}$\textbf{0.00} & 
            \textbf{1.00}$\bm{\pm}$\textbf{0.00} &
            \textbf{1.00}$\bm{\pm}$\textbf{0.00} \\

            &
            Score &
            0.38$\pm$0.43 & 
            0.13$\pm$0.24 & 
            0.76$\pm$0.35 & 
            \textbf{0.84}$\bm{\pm}$\textbf{0.31} &
            0.46$\pm$0.42 & 
            0.55$\pm$0.38 & 
            0.84$\pm$0.31 &
            \textbf{0.89}$\bm{\pm}$\textbf{0.21} \\

            &
            S-Rate &
            0.30$\pm$0.47 & 
            0.05$\pm$0.22 & 
            0.65$\pm$0.49 & 
            \textbf{0.75}$\bm{\pm}$\textbf{0.44} &
            0.35$\pm$0.49 & 
            0.40$\pm$0.50 & 
            \textbf{0.75}$\bm{\pm}$\textbf{0.44} &
            \textbf{0.75}$\bm{\pm}$\textbf{0.44} \\

            &
            SS-Rate &
            0.00$\pm$0.00 & 
            0.05$\pm$0.22 & 
            0.65$\pm$0.49 & 
            \textbf{0.75}$\bm{\pm}$\textbf{0.44} &
            0.00$\pm$0.00 & 
            0.40$\pm$0.50 & 
            \textbf{0.75}$\bm{\pm}$\textbf{0.44} &
            \textbf{0.75}$\bm{\pm}$\textbf{0.44} \\

            \specialrule{0.10em}{0pt}{0pt}
            
            \multirow{4}{*}{\emph{Task 2 - Close the teacup drawer}} &
            Safety &
            0.00$\pm$0.00 & 
            \textbf{1.00}$\bm{\pm}$\textbf{0.00} & 
            \textbf{1.00}$\bm{\pm}$\textbf{0.00} &
            \textbf{1.00}$\bm{\pm}$\textbf{0.00} & 
            0.00$\pm$0.00 & 
            \textbf{1.00}$\bm{\pm}$\textbf{0.00} & 
            \textbf{1.00}$\bm{\pm}$\textbf{0.00} &
            \textbf{1.00}$\bm{\pm}$\textbf{0.00} \\

            &
            Score &
            0.73$\pm$0.26 & 
            0.33$\pm$0.29 & 
            0.88$\pm$0.22 &
            \textbf{0.95}$\bm{\pm}$\textbf{0.15} &
            0.85$\pm$0.24 & 
            0.70$\pm$0.25 & 
            0.85$\pm$0.24 &
            \textbf{0.95}$\bm{\pm}$\textbf{0.15} \\

            &
            S-Rate &
            0.45$\pm$0.51 & 
            0.05$\pm$0.22 & 
            0.75$\pm$0.44 &
            \textbf{0.90}$\bm{\pm}$\textbf{0.31} &
            0.70$\pm$0.47 & 
            0.40$\pm$0.50 & 
            0.70$\pm$0.47 &
            \textbf{0.90}$\bm{\pm}$\textbf{0.31} \\

            &
            SS-Rate &
            0.00$\pm$0.00 & 
            0.05$\pm$0.22 & 
            0.75$\pm$0.44 &
            \textbf{0.90}$\bm{\pm}$\textbf{0.31} &
            0.00$\pm$0.00 & 
            0.40$\pm$0.50 & 
            0.70$\pm$0.47 &
            \textbf{0.90}$\bm{\pm}$\textbf{0.31} \\

            \specialrule{0.1em}{1pt}{1pt}
            \specialrule{0.1em}{1pt}{1pt}
            
        \end{tabular}%

\end{table*}
\begin{figure*}[t]
    \centering
    \captionsetup[subfigure]{skip=2pt,font=irvlasubplot,justification=centering}

    \begin{subfigure}[t]{0.275\textwidth}
        \centering
        \vspace{0pt}

        \includegraphics[
            width=\linewidth,
            height=0.455\textwidth,
        ]{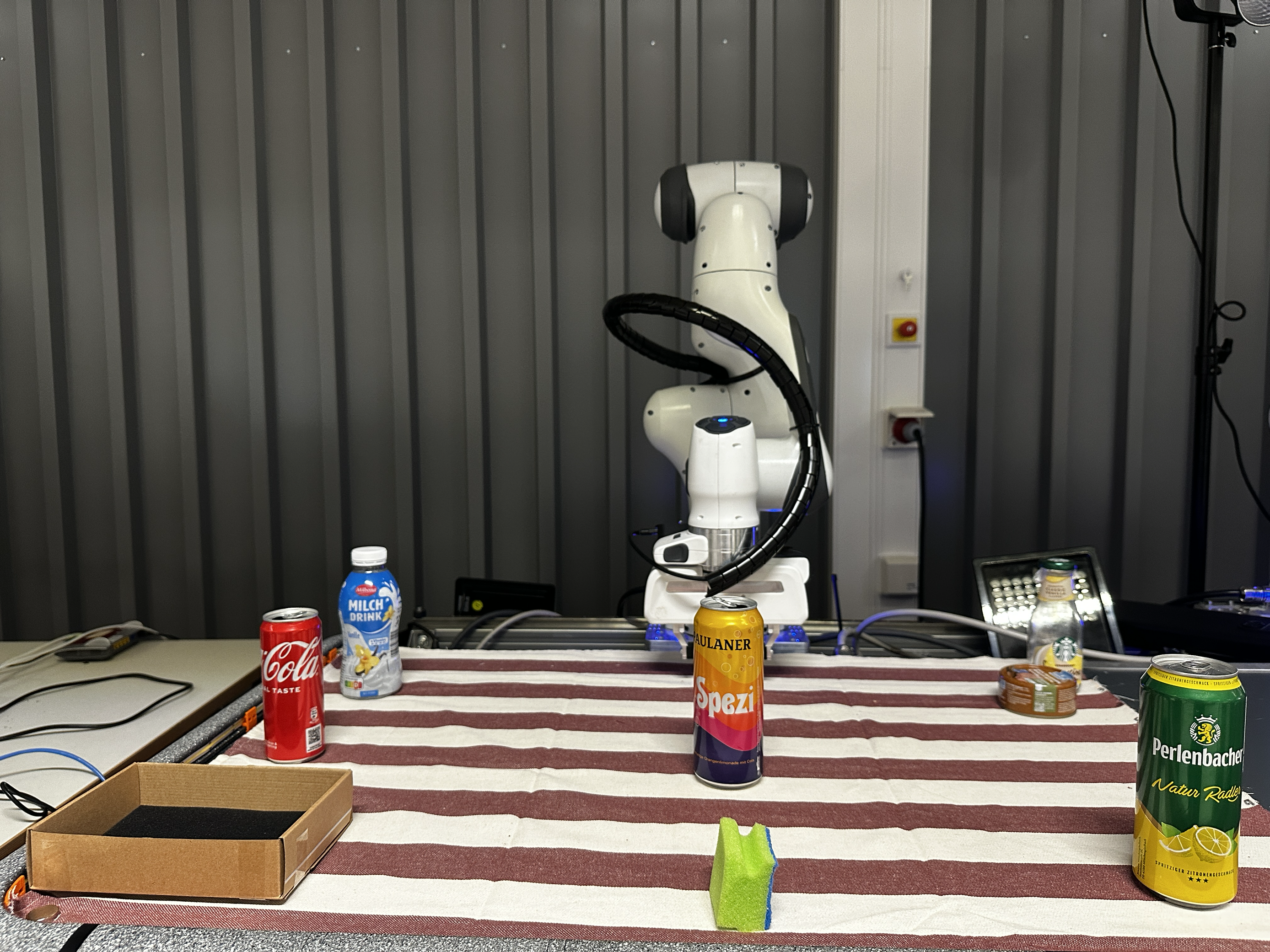}

        \vspace{0.2em}

        \includegraphics[
            width=\linewidth,
            height=0.455\textwidth,
        ]{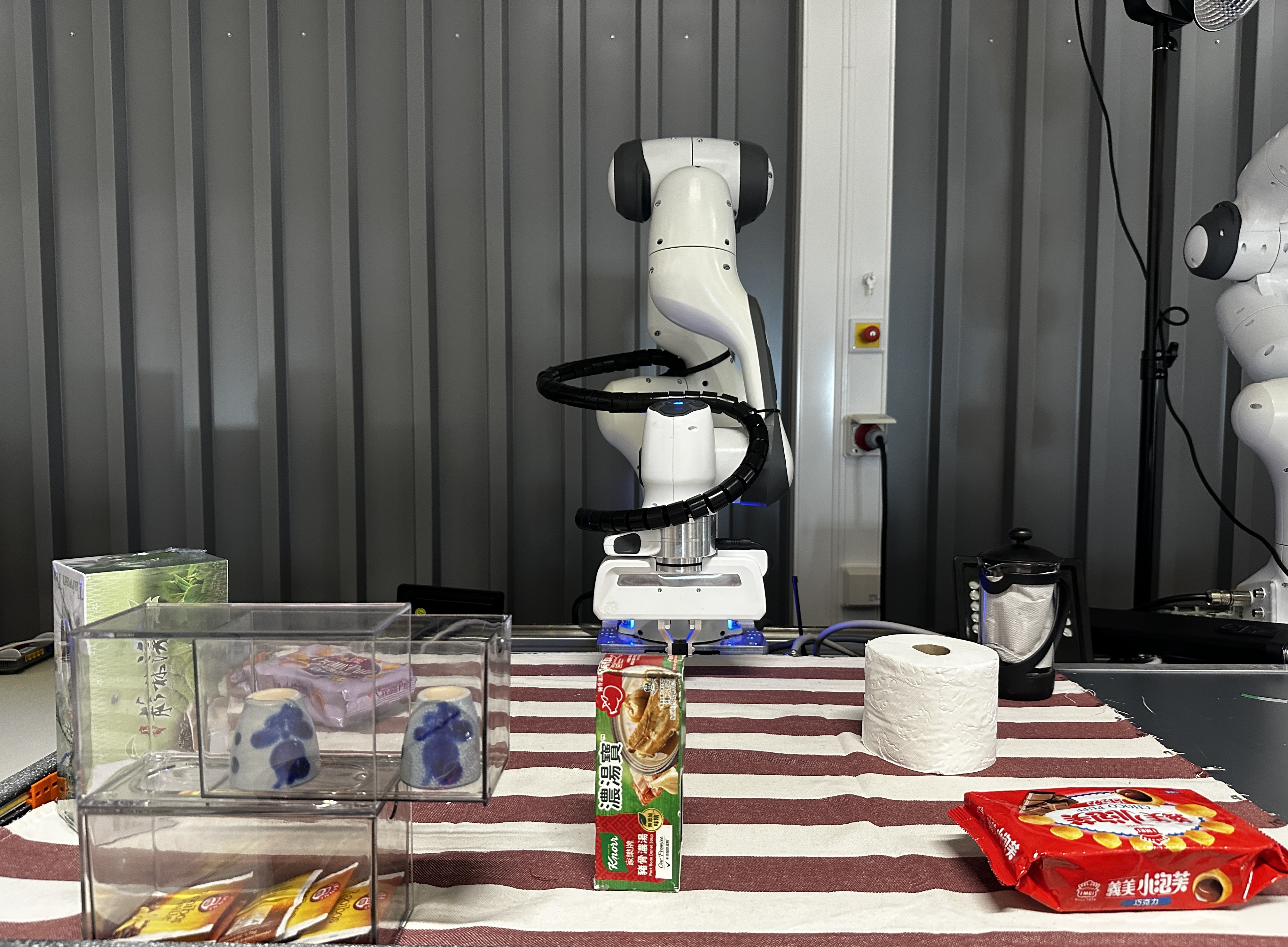}

        \caption{Tasks for evaluation}
        \label{fig:task_type}
    \end{subfigure}%
    \hspace{0.005\textwidth}%
    \begin{subfigure}[t]{0.235\textwidth}
        \centering
        \vspace{0pt}

        \includegraphics[
            width=\linewidth,
            height=4.5cm,
        ]{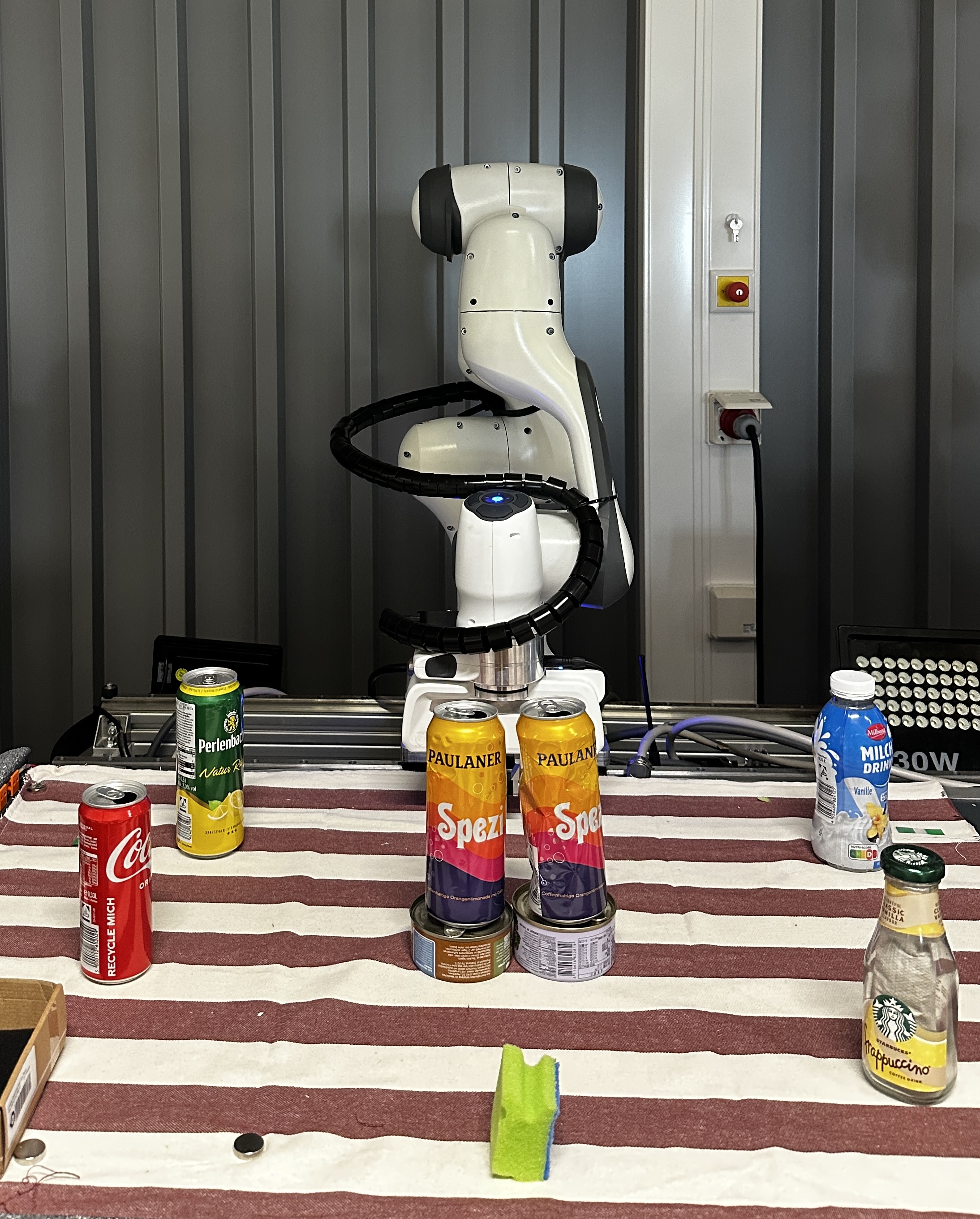}

        \caption{OOD for obstacle size.}
        \label{fig:general_size}
    \end{subfigure}%
    \hspace{0.005\textwidth}%
    \begin{subfigure}[t]{0.235\textwidth}
        \centering
        \vspace{0pt}

        \includegraphics[
            width=\linewidth,
            height=4.5cm,
        ]{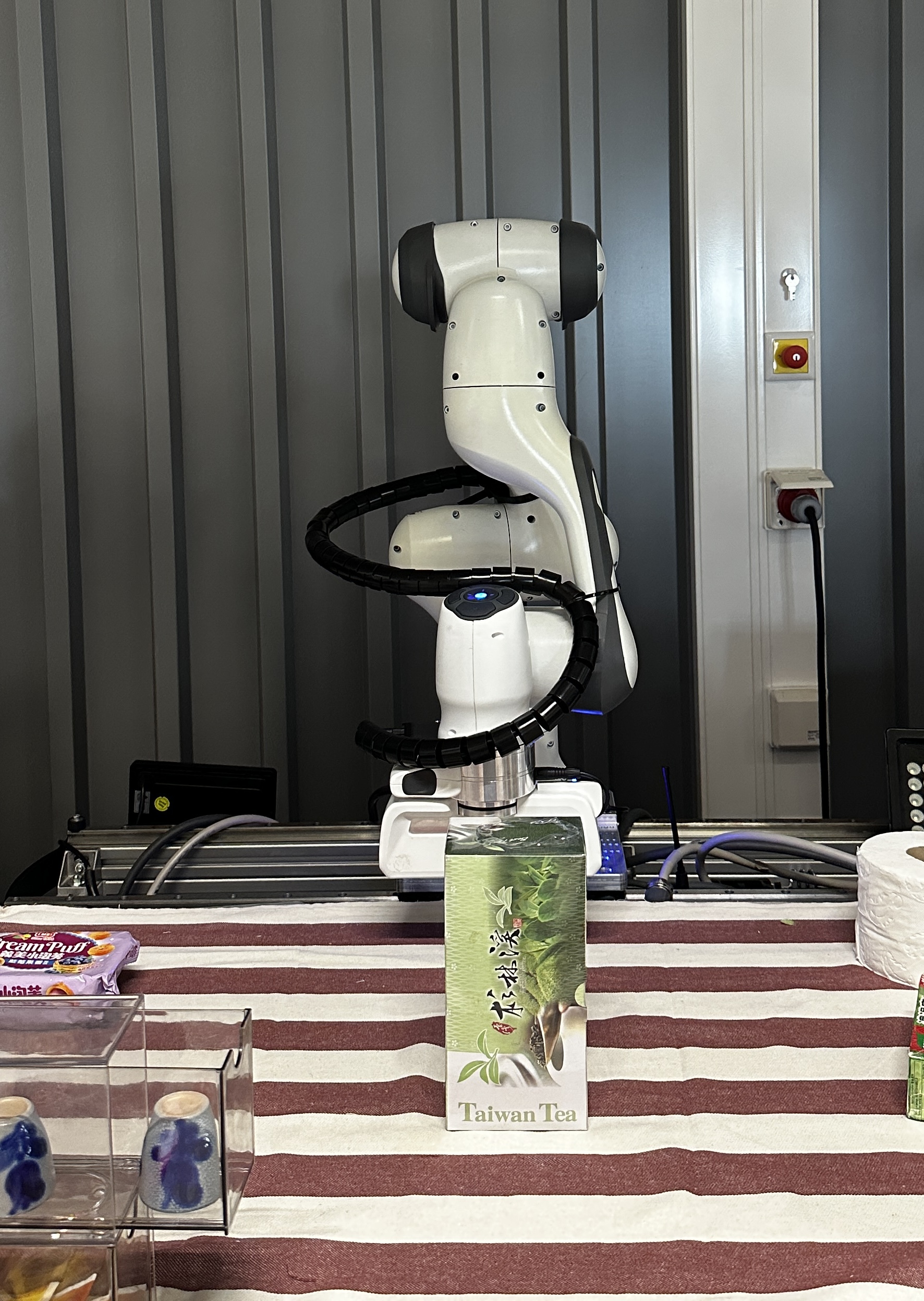}

        \caption{OOD for obstacle type.}
        \label{fig:general_type}
    \end{subfigure}%
    \hspace{0.005\textwidth}%
    \begin{subfigure}[t]{0.235\textwidth}
        \centering
        \vspace{0pt}

        \includegraphics[
            width=\linewidth,
            height=4.5cm,
        ]{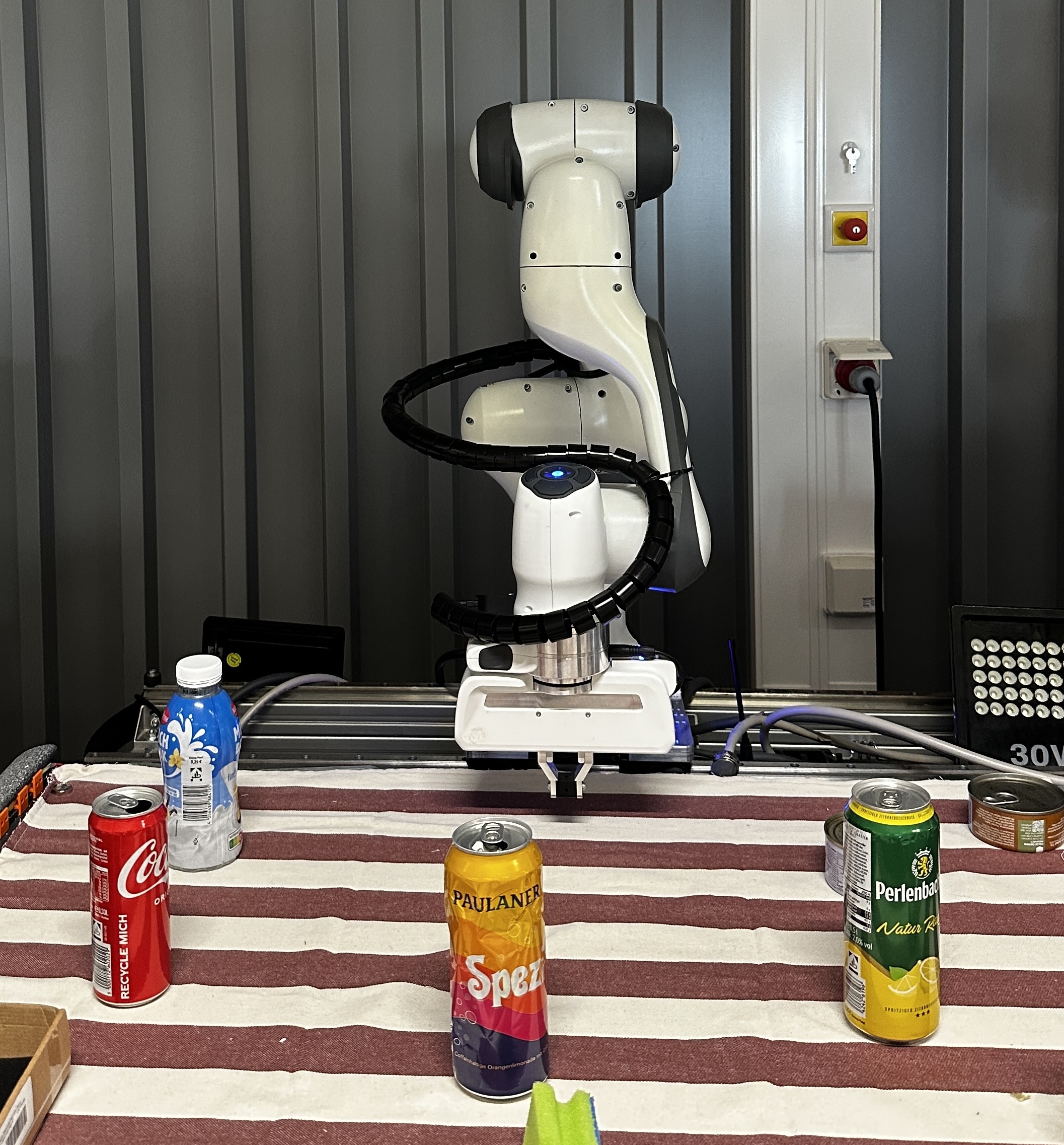}

        \caption{OOD for obstacle location.}
        \label{fig:general_location}
    \end{subfigure}
\caption{\textbf{Evaluation setup.} \Cref{fig:task_type}: \emph{Task 1: Put sponge into basket} (top) and \emph{Task 2: Close the teacup drawer} (bottom), with the obstacle on the nominal path to the goal. \Cref{fig:general_size,fig:general_type,fig:general_location}: the three OOD settings, each varying obstacle size, type, or location relative to the data-collection configurations.}
    \label{fig:general_layout}
\vspace{-0.5em}
\end{figure*}

\begingroup
\makeatletter
\@secpenalty=10000
\makeatother
\paragraph{Tasks}
We consider two language-specified tabletop manipulation tasks,
visualized in~\Cref{fig:general_layout}~(a), requiring successful
completion without colliding with obstacles:
\par
\endgroup
\begin{itemize}
    \item \emph{Task 1 - Put sponge into basket.}
    The robot approaches and grasps a sponge placed on the table, transports it to the basket, and places it inside.

    \item \emph{Task 2 - Close the teacup drawer.}
    The robot approaches an initially open drawer, makes contact with the drawer front, and pushes it until the drawer is closed.
\end{itemize}
\paragraph{Safety Constraint}
For each task, an obstacle is placed along the nominal shortest path to the task goal, requiring the robot to deviate from its direct motion to avoid collision. We use GroundingDINO~\cite{liu2024grounding} to localize the obstacle in the RGB image and obtain its bounding box. The corresponding depth measurements from the RGB-D camera are then used to reconstruct the obstacle point cloud, which is approximated by an ellipsoid. The robot end effector is similarly represented by an ellipsoid. Following~\cite{hu2025vlsa}, the safety constraint is defined through the separation between these two ellipsoids, and the safety filter modifies the commanded motion to prevent their intersection.

\paragraph{Evaluation Metrics}
We evaluate each method using the following metrics calculated from 20 rollouts and report the mean and standard deviation:
\begin{itemize}
    \item \textbf{Safety:}
    The fraction of episodes that remain collision-free throughout the entire execution.
    
    \item \textbf{Score:}
    A progress score in $[0,1]$ that measures partial task completion through predefined task milestones. For \emph{Put sponge into basket}, the milestones are: (i) approach the sponge, (ii) grasp the sponge, (iii) move the sponge to the basket, and (iv) place it inside the basket. For \emph{Close the teacup drawer}, the milestones are: (i) approach the drawer, (ii) make contact with the drawer, and (iii) push the drawer until closed.

    \item \textbf{Success Rate (S-Rate):}
    The fraction of episodes in which the complete task is successfully accomplished.

    \item \textbf{Safe Success Rate (SS-Rate):}
    The fraction of episodes in which the task is successfully completed while remaining collision-free throughout the entire execution.
\end{itemize}

\paragraph{VLA Policies}
We evaluate two representative VLA models, $\pi_{0.5}$~\cite{intelligence2025pi} and GR00T N1.5~\cite{bjorck2025gr00t}. The policy observation consists of robot joint states together with RGB images from the wrist-mounted and third-person cameras. The action specifies the desired Cartesian end effector displacement and gripper width. For each task, we first collect $50$ nominal demonstrations as a base dataset shared for all VLAs.

\begin{figure*}[t]
\centering
\begin{tikzpicture}
\def\oodbars#1{
  \foreach \pre/\off in {Pi/0.33, Gr/1.45}{
    \edef\next{%
    \noexpand\addplot[irvla bar=strat_vla_normal, bar shift=-0.33]
      table[x expr=\noexpand\coordindex*2.52+\off, y=\pre NormMean] {#1};
    \noexpand\addplot[irvla bar=strat_sf_normal, bar shift=-0.11]
      table[x expr=\noexpand\coordindex*2.52+\off, y=\pre NormFMean] {#1};
    \noexpand\addplot[irvla bar=strat_sf_diverse, bar shift=0.11]
      table[x expr=\noexpand\coordindex*2.52+\off, y=\pre DivFMean] {#1};
    \noexpand\addplot[irvla bar=strat_sf_multi, bar shift=0.33]
      table[x expr=\noexpand\coordindex*2.52+\off, y=\pre MultiFMean] {#1};}\next}
  \foreach \x in {0.89, 3.41, 5.93}{
    \edef\next{\noexpand\draw[fig_rule, line width=0.5pt, dash pattern=on 2pt off 2pt]
      (axis cs:\x,0) -- (axis cs:\x,1);}\next}}
\begin{groupplot}[
    group style={group size=1 by 2, vertical sep=0.34in},
    irvla, ybar, /pgf/bar width=0.20,
    width=\dimexpr\textwidth-0.50in\relax, height=0.75in,
    xmin=-0.22, xmax=7.04,
    xtick={0.33, 1.45, 2.85, 3.97, 5.37, 6.49},
    xticklabels={\figpizero, GR00T, \figpizero, GR00T, \figpizero, GR00T},
    xtick style={draw=none}, xticklabel style={yshift=1pt},
    ylabel={SS-Rate}, ylabel style={yshift=-2pt}, xlabel style={yshift=3pt},
    point meta=rawy,                          
    y filter/.expression={y==0 ? 0.03 : y},   
    irvla value,
    every node near coord/.append style={fill=white},  
    legend image code/.code={\draw[#1] (0cm,-0.09cm) rectangle (0.30cm,0.10cm);},
  ]
  \nextgroupplot[xlabel={Task 1: Put sponge into basket},
    legend to name=oodlegend, legend columns=4,
    legend style={/tikz/every even column/.append style={column sep=1.3em}}]
  \oodbars{figures/data/task1_ood.txt}
  \legend{VLA (Normal), VLA-SF (Normal), VLA-SF (Diverse), VLA-SF (Multi)}
  \foreach \x/\n in {0.89/OOD Size, 3.41/OOD Type, 5.93/OOD Location}{
    \edef\next{\noexpand\node[font=\noexpand\figfont{8.5}, text=fig_ink, anchor=base]
      at (axis cs:\x,1) [yshift=0.06in] {\n};}\next}
  \nextgroupplot[xlabel={Task 2: Close the teacup drawer}]
  \oodbars{figures/data/task2_ood.txt}
\end{groupplot}
\node[anchor=south] at ([yshift=0.11in]group c1r1.north) {\pgfplotslegendfromname{oodlegend}};
\end{tikzpicture}%
\caption{
    \textbf{OOD evaluation.} Mean SS-Rate over $20$ rollouts across
    all strategies with $\pi_{0.5}$ and GR00T N1.5 under three OOD
    shifts (obstacle size, type, location) for
    \emph{Task 1: Put sponge into basket} (top) and
    \emph{Task 2: Close the teacup drawer} (bottom).
    Both InteR-VLA strategies consistently outperform nominal-data baselines across all OOD shifts.
}
    \label{fig:ood_ssr_task12}
\end{figure*}
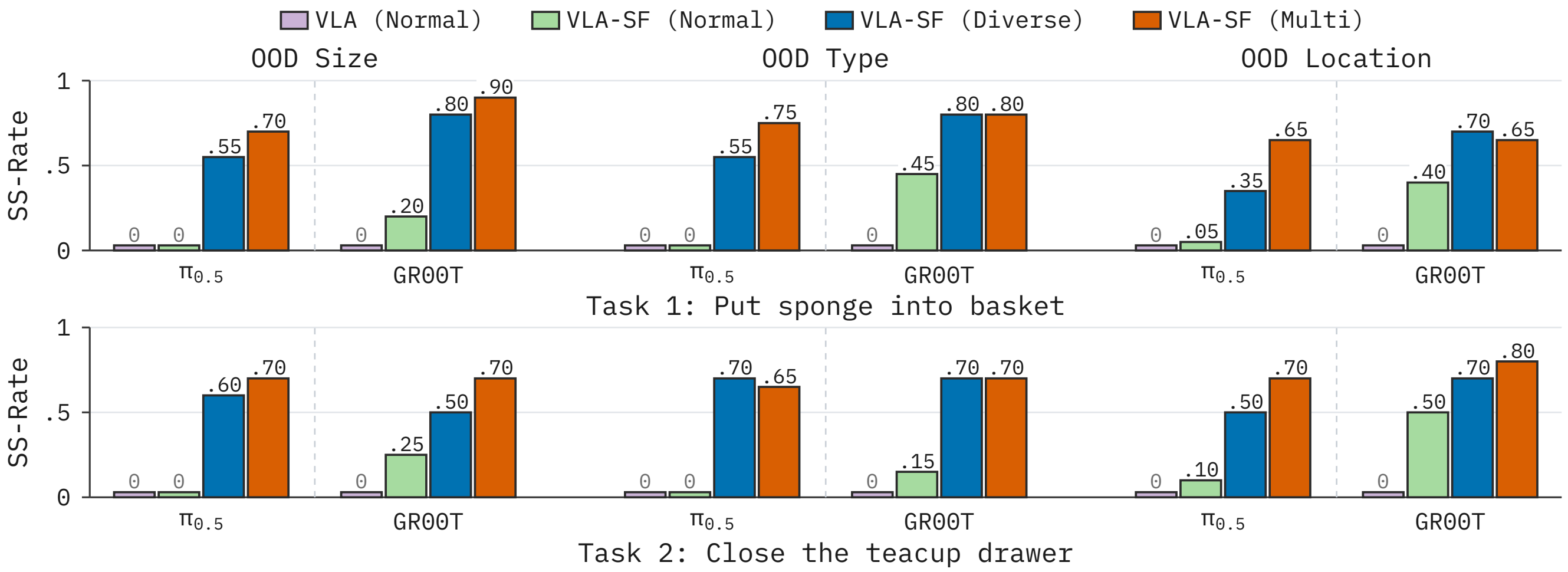

\paragraph{Data-Collection Strategies}
We further collect 150 demonstrations beyond the base dataset for each of the following settings. The first two are nominal-data baselines; the last two are the InteR-VLA strategies of \cref{sec:deployment_distribution}:
\begin{itemize}
    \item \textbf{VLA (Normal):}
    The VLA policy is fine-tuned using nominal demonstrations and directly deployed without a safety filter. During data collection, obstacles are present in the environment but do not block the nominal task path; therefore, the demonstrator completes the task directly without explicit collision-avoidance maneuvers. This tests whether nominal demonstrations alone teach the policy to avoid a blocking obstacle.

    \item \textbf{VLA-SF (Normal):}
    The same policy and nominal demonstrations as \textbf{VLA (Normal)} are used, while a safety filter is additionally applied during deployment, following the implementation in VLSA~\cite{hu2025vlsa}. This tests whether the safety filter alone yields safe task success without changing the data.

    \item \textbf{VLA-SF (Diverse):}
    Obstacle-free diverse demonstrations (see \Cref{sec:data_collection}), e.g., curved, left-offset, and right-offset motions, with the safety filter at deployment.

    \item \textbf{VLA-SF (Multi):}
    Obstacle-aware multimodal demonstrations (see \Cref{sec:data_collection}), e.g., above, left, and right avoidance, with the safety filter at deployment.

\end{itemize}
\begin{figure}[t]
    \centering
    \includegraphics{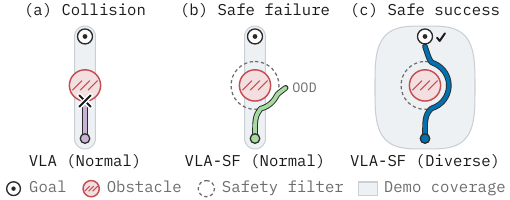}%
    \caption{\textbf{Intervention robustness schematic.} A nominal-data policy collides unfiltered~(a). Filtering (dashed: safe-set boundary) pushes it beyond nominal coverage, causing failure~(b); broader coverage contains the corrected path~(c).}
    \label{fig:main_traj}
\end{figure}%

\subsection{Real-World Experiment Results}

\paragraph{In-distribution (ID) Obstacle Evaluation}
We first evaluate all strategies under ID deployment conditions across both manipulation tasks and both VLA models. The obstacle type, size, and placement are sampled from the same distribution used for data collection, and in every episode the obstacle blocks the nominal shortest path to the goal. This setting evaluates whether different data-collection strategies and safety-filter integration can improve safe task execution when the environment remains within the training distribution under the same demonstration budget.

As shown in \Cref{tab:main_exp}, \textbf{VLA-SF (Multi)} achieves the best overall performance across both VLA models and tasks, with consistently high success and safety. \textbf{VLA-SF (Diverse)} also outperforms VLA-SF (Normal) in SS-Rate on both tasks and models, even without obstacle-specific data collection. For Task~1 with GR00T~N1.5, VLA-SF (Diverse) matches VLA-SF (Multi) at $0.75$ SS-Rate, showing that deployment-relevant coverage can arise from naturally collected diverse demonstrations without designing obstacle-specific scenarios. The observed trend is consistent with \Cref{th:sda}: strategies designed to better align demonstration coverage with safety-filtered deployment show higher SS-Rate than nominal-data training. In contrast, \textbf{VLA (Normal)} collides in every episode despite seeing the same obstacle types during training, showing that collision-free nominal demonstrations with non-blocking obstacles do not teach avoidance behavior. \textbf{VLA-SF (Normal)} improves safety but sacrifices task success because filter corrections drive the robot toward states poorly covered by nominal demonstrations. \Cref{fig:main_traj} schematically summarizes this mechanism: nominal-data policies may either collide without filtering or fail after safety correction, whereas broader demonstration coverage can support safe task completion.\looseness=-1

\paragraph{Out-of-distribution (OOD) Obstacle Evaluation}
We evaluate intervention robustness under OOD obstacle configurations across both tasks and VLA models. We consider distribution shifts from data collection along three dimensions, which are visualized in~\Cref{fig:general_layout}-(b) to~(d):

\textbf{Obstacle Size.}
We evaluate three unseen obstacle geometries: $2\!\times\!1$ the nominal width, $1.3\!\times\!1$ the height, and their combination. The height scaling is limited to $1.3\!\times\!1$ to leave sufficient free space for collision-free task execution.

\textbf{Obstacle Type.}
We replace the obstacles with unseen types to test generalization across appearance and geometry.

\textbf{Obstacle Location.}
We place obstacles at unseen locations outside the training region that block the nominal path while preserving a collision-free solution, testing spatial generalization beyond observed placements.

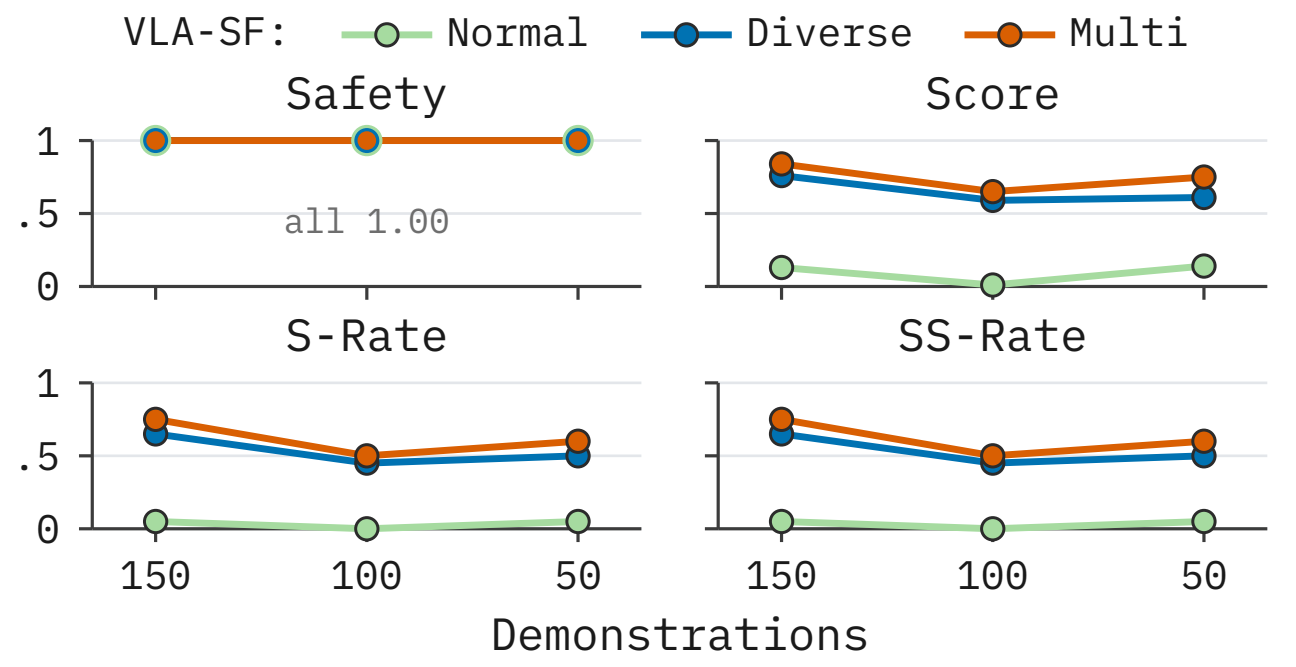
\begin{figure}[t]
\centering
\begin{tikzpicture}
\def\dsplot#1{
  \addplot[irvla line=strat_sf_normal]
    table[x expr=50*\thisrow{datasize}, y=normal_filter] {figures/data/#1};
  \addplot[irvla line=strat_sf_diverse]
    table[x expr=50*\thisrow{datasize}, y=diverse] {figures/data/#1};
  \addplot[irvla line=strat_sf_multi]
    table[x expr=50*\thisrow{datasize}, y=multi] {figures/data/#1};}
\begin{groupplot}[
    group style={group size=2 by 2, horizontal sep=0.20in, vertical sep=0.251in,
                 xticklabels at=edge bottom, yticklabels at=edge left},
    irvla,
    width=\dimexpr(\columnwidth-0.54in)/2\relax, height=0.38in,
    x dir=reverse, xmin=35, xmax=165, restrict x to domain=40:160,
    xtick={50,100,150}, title style={anchor=base},  
  ]
  \nextgroupplot[title={Safety}]
  \addplot[irvla line=strat_sf_normal, mark size=2.8pt, mark options={line width=0.3pt}]
    table[x expr=50*\thisrow{datasize}, y=normal_filter] {figures/data/t1_datasize_safe.txt};
  \addplot[irvla line=strat_sf_diverse, mark size=2.2pt, mark options={line width=0.3pt}]
    table[x expr=50*\thisrow{datasize}, y=diverse] {figures/data/t1_datasize_safe.txt};
  \addplot[irvla line=strat_sf_multi, mark size=1.6pt, mark options={line width=0.3pt}]
    table[x expr=50*\thisrow{datasize}, y=multi] {figures/data/t1_datasize_safe.txt};
  \node[font=\figfont{6.5}, text=fig_muted] at (axis cs:100,0.45) {all 1.00};
  \nextgroupplot[title={Score},
    legend to name=dslegend, legend columns=4,
    legend style={/tikz/every even column/.append style={column sep=0.7em}}]
  \addlegendimage{empty legend}\addlegendentry{\hspace{-1.2em}VLA-SF:}
  \dsplot{t1_datasize_score.txt}
  \addlegendentry{Normal}\addlegendentry{Diverse}\addlegendentry{Multi}
  \nextgroupplot[title={S-Rate}]
  \dsplot{t1_datasize_sr.txt}
  
  \nextgroupplot[title={SS-Rate}]
  \dsplot{t1_datasize_ssr.txt}
\end{groupplot}
\path (group c1r2.south) -- node[midway, anchor=north, yshift=-0.18in,
  font=\figfont{8.5}, text=fig_ink] {Demonstrations} (group c2r2.south);
\path (group c1r1.north) -- node[midway, anchor=south, yshift=0.122in]
  {\pgfplotslegendfromname{dslegend}} (group c2r1.north);
\end{tikzpicture}%
\caption{\textbf{Effect of reducing demonstration budget (Task~1).} Each panel plots one metric across budget for VLA $\pi_{0.5}$. Both InteR-VLA strategies degrade but keep a clear SS-Rate advantage over \textbf{VLA-SF (Normal)} at the smallest budget.}
    \label{fig:abla_data_size}
\end{figure}

\begin{figure}[!b]
\centering
\begin{tikzpicture}
\begin{groupplot}[
    group style={group size=2 by 1, horizontal sep=0.16in},
    irvla, height=0.48in, xlabel style={yshift=2pt},
  ]
  \nextgroupplot[width=\dimexpr(\columnwidth-0.66in)*56/100\relax,
    xmin=0.7, xmax=3.3, xtick={1,2,3}, ylabel={SS-Rate}, ylabel style={yshift=-2pt},
    xlabel={(a) Avoidance modes},
    legend style={at={(0.97,0.04)}, anchor=south east}, legend cell align=left]
  \addplot[irvla line=strat_sf_multi, nodes near coords, point meta=y,
    every node near coord/.append style={font=\figfont{6.5}, text=fig_ink,
      anchor=base, yshift=4pt, inner sep=0pt,
      /pgf/number format/.cd, fixed, fixed zerofill, precision=2, skip 0.,
      assume math mode=true}]
    table[x=mod, y=multi] {figures/data/t1_modality.txt};
  \addlegendentry{VLA-SF (Multi)}
  \nextgroupplot[width=\dimexpr(\columnwidth-0.66in)*44/100\relax,
    ybar, /pgf/bar width=0.52, xmin=-0.6, xmax=1.6, xtick={0,1},
    xticklabels={Normal, Mixed}, xtick style={draw=none}, yticklabels={},
    xlabel={(b) Task transfer},
    visualization depends on={\thisrow{std} \as \sd},
    nodes near coords={\pgfmathprintnumber{\pgfplotspointmeta}\textpm%
      \pgfmathprintnumber{\sd}},
    every node near coord/.append style={font=\figfont{6.5}, text=fig_ink,
      anchor=base, yshift=1.2pt, inner sep=0pt,
      /pgf/number format/.cd, fixed, fixed zerofill, precision=2, skip 0.,
      assume math mode=true}]
  \addplot[irvla bar=strat_sf_normal, bar shift=0pt] table[x=x, y=mean] {
    x mean std
    0 0.05 0.22
  };
  \addplot[irvla bar=strat_sf_multi, bar shift=0pt] table[x=x, y=mean] {
    x mean std
    1 0.65 0.49
  };
\end{groupplot}
\end{tikzpicture}%
\caption{
        \textbf{Avoidance modes and task transfer} (SS-Rate, equal dataset sizes). (a) Task~1: SS-Rate rises with the number of avoidance modes. (b) Task~2 with $\pi_{0.5}$: mixing 100 Task~1 Multi demonstrations with 100 Task~2 nominal demonstrations (Mixed) reaches $0.65$, compared with $0.05$ for only mixing nominal data (Normal).
    }\label{fig:abla_modality_noobs_task_transfer}
\vspace{-0.4em}
\end{figure}
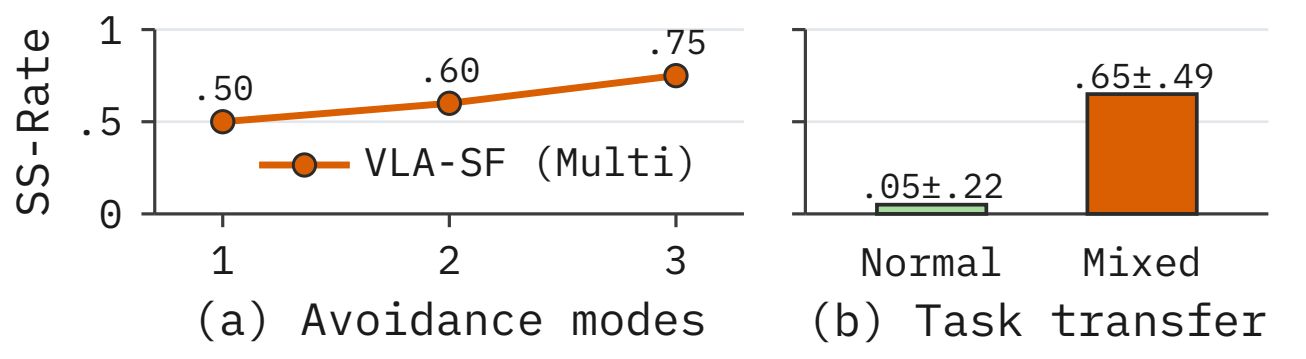

As shown in \Cref{fig:ood_ssr_task12}, \textbf{VLA-SF (Multi)} and \textbf{VLA-SF (Diverse)} consistently outperform nominal-data baselines across tasks, VLA models, and OOD settings under the same data-collection budget. GR00T N1.5 generally performs better than $\pi_{0.5}$, with \textbf{VLA-SF (Multi)} reaching up to $0.90$ SS-Rate under OOD obstacle sizes. \textbf{VLA-SF (Diverse)} remains competitive with, and sometimes surpasses, VLA-SF (Multi), e.g., $0.70$ vs.\ $0.65$ under OOD obstacle locations in Task~1 with GR00T~N1.5, despite training without obstacles. This shows that diverse demonstrations can provide sufficient task-relevant coverage to preserve competence under strong safety-filter corrections without obstacle-specific data. All safety-filtered methods satisfy the safety constraints, whereas \textbf{VLA-SF (Normal)} underperforms both InteR-VLA strategies by at least $0.20$ SS-Rate in every OOD setting because nominal demonstrations poorly cover post-intervention states. Without the safety filter, \textbf{VLA (Normal)} achieves zero SS-Rate across OOD settings due to collisions. Overall, diverse demonstration coverage improves the robustness of interventions to unseen obstacle configurations, enabling safe and successful task completion with safety filters.

\subsection{Ablation Study}
We conduct a series of ablation studies to better understand how different factors affect safe task execution.

\paragraph{Effect of Reduced Dataset Size}
We evaluate reduced demonstration budgets. As shown in \Cref{fig:abla_data_size}, \textbf{VLA-SF (Multi)} and \textbf{VLA-SF (Diverse)} perform worse as datasets shrink. Yet, with only 50 further demonstrations, both still exceed VLA-SF (Normal) in SS-Rate ($0.60$ $\&$ $0.50$ vs.\ $\!0.05$)\looseness=-1.

\begin{table}[t]
    \vspace{4pt}
    \centering
\caption{\textbf{Safety without the filter (Task~1).} $\pi_{0.5}$ policies trained with our data strategies are deployed without the filter (VLA) and with it (VLA-SF). Safety decreases without filtering, especially under OOD conditions, showing that our data cannot replace test-time filtering.}
    \label{tab:abla_no_sf}
    \footnotesize
    \renewcommand{\arraystretch}{1.2}
    \setlength{\tabcolsep}{4pt}
    \begin{tabular}{c|cc|cc}
        \specialrule{0.12em}{1pt}{1pt}
        \multirow{2}{*}{\textbf{Safety}}
        & \multicolumn{2}{c|}{\textbf{w/o filter:} VLA}
        & \multicolumn{2}{c}{\textbf{w/ filter:} VLA-SF} \\
        \cmidrule(lr){2-3} \cmidrule(lr){4-5}
        & Diverse & Multi & Diverse & Multi \\
        \specialrule{0.10em}{0pt}{0pt}
        ID case  & 0.80 & 0.65 & 1.00 & 1.00 \\
        OOD case & 0.50 & 0.35 & 1.00 & 1.00 \\
        \specialrule{0.1em}{1pt}{1pt}
    \end{tabular}
\vspace{-0.3em}
\end{table}

\begin{figure}[b]
    \centering
    \begin{minipage}[b]{0.68\linewidth}
        \centering
        \IfFileExists{figures/data/unseen_object_layout.pdf}{%
            \includegraphics[width=\linewidth, height=2.6cm,trim=0 0 0 28bp,clip]{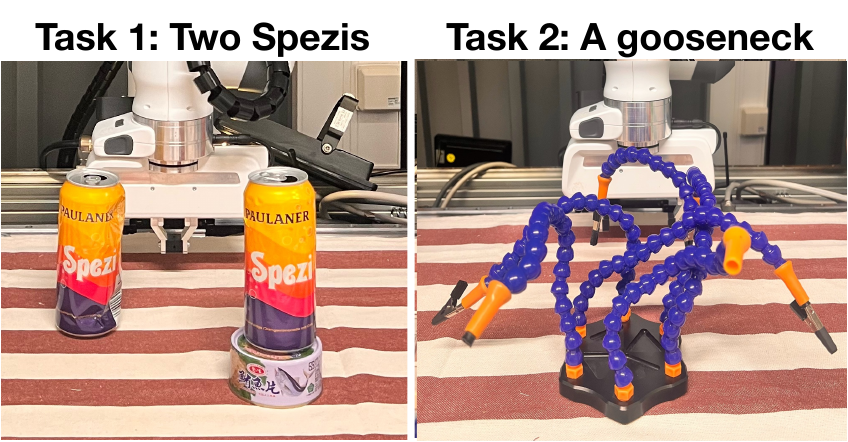}%
        }{%
            \fbox{\parbox[c][0.45\linewidth][c]{0.88\linewidth}{\centering\tiny Layout PDF placeholder}}%
        }\par
        \nointerlineskip\vspace{2.5pt}%
        \begin{minipage}[t]{0.4828\linewidth}
            \centering\figfont{6.5}Task 1: Two cans
        \end{minipage}%
        \hspace{0.0049\linewidth}%
        \begin{minipage}[t]{0.5123\linewidth}
            \centering\figfont{6.5}Task 2: A gooseneck
        \end{minipage}
    \end{minipage}%
      \hspace*{-2pt}%
    \begin{minipage}[b]{0.32\linewidth}
        \raggedleft
        \begin{tikzpicture}[baseline=(xl.base)]
        \begin{axis}[
            irvla, xbar, /pgf/bar width=6pt,
            xmin=0, xmax=1, xtick={0,0.5,1},
            ymajorgrids=false, xmajorgrids=false,
            xticklabel={\pgfmathprintnumber[assume math mode=true, skip 0.]{\tick}},
            xticklabel style={anchor=base, yshift=-7pt},
            ytick style={draw=none}, yticklabel style={xshift=1pt},
            axis y line*=left, axis x line*=bottom,
            width=0.50in, height=1.002in, ymin=-0.6, ymax=7.55,
            ytick={0,1,2,4.2,5.2,6.2}, yticklabels={Multi, Diverse, Normal, Multi, Diverse, Normal},
            point meta=rawx, x filter/.expression={x==0 ? 0.03 : x},
            nodes near coords={\pgfmathfloatifflags{\pgfplotspointmeta}{0}%
              {\textcolor{fig_muted}{0}}%
              {\pgfmathprintnumber[fixed, fixed zerofill, precision=2, skip 0.,
                                   assume math mode=true]{\pgfplotspointmeta}}},
            every node near coord/.append style={font=\figfont{6.5}, text=fig_ink,
              anchor=west, xshift=1.2pt, inner sep=0pt},
          ]
          \foreach \gx in {0.5,1} {
            \edef\temp{\noexpand\draw[fig_grid, line width=0.5pt]
              (axis cs:\gx,-0.6) -- (axis cs:\gx,2.58)
              (axis cs:\gx,3.6) -- (axis cs:\gx,6.78);}
            \temp
          }
          \draw[line width=0.6pt, fig_ink!85] (axis cs:0,3.6) -- (axis cs:1,3.6);
          \addplot[irvla bar=strat_sf_normal,  bar shift=0pt] coordinates {(0.00,6.2) (0.05,2)};
          \addplot[irvla bar=strat_sf_diverse, bar shift=0pt] coordinates {(0.55,5.2) (0.55,1)};
          \addplot[irvla bar=strat_sf_multi,   bar shift=0pt] coordinates {(0.65,4.2) (0.55,0)};
          \node[font=\figfont{6.5}, text=fig_muted, anchor=base west, inner sep=0pt]
            at ([xshift=2pt]axis cs:0,7.0) {Task 1};
          \draw[fig_rule, line width=0.5pt]
            ([xshift=-37pt]axis cs:0,6.78) -- (axis cs:1,6.78);
          \node[font=\figfont{6.5}, text=fig_muted, anchor=base west, inner sep=0pt]
            at ([xshift=2pt]axis cs:0,2.8) {Task 2};
          \draw[fig_rule, line width=0.5pt]
            ([xshift=-37pt]axis cs:0,2.58) -- (axis cs:1,2.58);
          \node[name=xl, font=\figfont{7.5}, text=fig_ink, anchor=base east]
            at ([xshift=-3.5pt, yshift=-7pt]axis cs:0,-0.6) {SS-Rate};
        \end{axis}
        \end{tikzpicture}%
      \end{minipage}\hspace*{4pt}
\caption{
    \textbf{Extreme OOD layouts.} Scenarios beyond the OOD settings of \Cref{fig:ood_ssr_task12}: two obstacles blocking the path in Task~1, and an unseen gooseneck obstacle in Task~2. SS-Rate of VLA-SF with $\pi_{0.5}$ (mean, $20$ rollouts, right) shows nominal demonstrations failing almost everywhere; both InteR-VLA strategies complete over half the episodes.
}\label{fig:abla_unseen_object}
\end{figure}
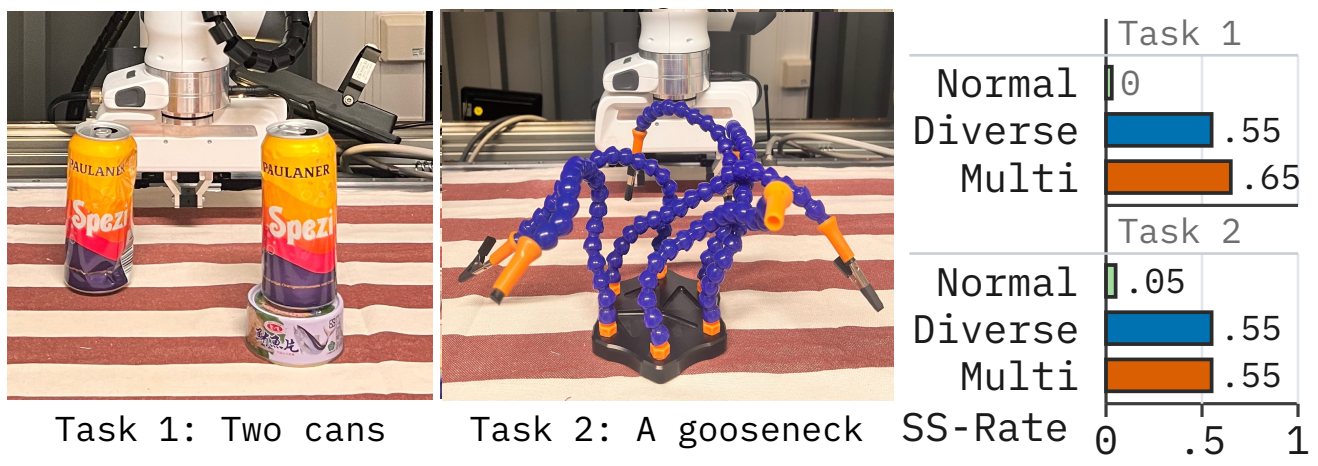%

\paragraph{Effect of Avoidance Modes}
We vary avoidance diversity during data collection using one mode (above), two modes (left and right), and three modes (above, left, and right) in Task~1. As shown in \Cref{fig:abla_modality_noobs_task_transfer}-(a), SS-Rate increases with more modes under the same data budget. Thus, the gain in intervention robustness stems from behavioral diversity rather than data quantity, consistent with our hypothesis that additional modes broaden task-relevant coverage and overlap more with states reached after safety-filter interventions.

\paragraph{Cross-Task Transfer}
To assess whether InteR-VLA can transfer intervention robustness across tasks, we train $\pi_{0.5}$ under the same data budget using 100 demonstrations (Multi) from Task~1 and 100 demonstrations (Normal) from Task~2, and evaluate on Task~2. As shown in \Cref{fig:abla_modality_noobs_task_transfer}-(b), the mixed dataset yields $0.65$ SS-Rate, compared with $0.05$ for nominal-only training, indicating that diverse demonstrations can improve intervention robustness across tasks.

\paragraph{Safety Without a Safety Filter}
To test whether our demonstrations alone ensure safe execution, we deploy $\pi_{0.5}$ policies trained with both strategies without filtering. In \Cref{tab:abla_no_sf}, safety drops from $1.00$ with filtering to $0.80$ (Diverse) and $0.65$ (Multi) in ID, and further to $0.50$ and $0.35$ in OOD. These results show that demonstrations alone are insufficient for safety, highlighting the need for InteR-VLA combining diverse demonstrations with test-time filtering.

\paragraph{Extreme OOD Evaluation}
We further evaluate InteR-VLA under extreme OOD scenarios beyond the single-obstacle variations considered above. As shown in \Cref{fig:abla_unseen_object}, Task~1 involves two simultaneous obstacles blocking the nominal path, whereas Task~2 uses an unseen object with substantially different geometry and visual appearance from the objects seen during data collection. InteR-VLA achieves at least $0.55$ SS-Rate in both settings, compared with $0$ and $0.05$ for nominal-data training, demonstrating intervention robustness under novel layouts, geometries, and visual shifts.
\section{Conclusion}

In this work, we introduced the InteR-VLA framework, which leverages diverse demonstrations to reduce the distribution mismatch induced by safety filters. Our approach separates the roles of training and deployment: safety filters enforce safety at test time, while diverse demonstrations broaden task-relevant coverage to preserve competence under intervention. Across two tabletop manipulation tasks and ID and OOD scenarios, InteR-VLA improves the Safe Success Rate from $0\%$ up to $90\%$. Under a fixed data budget, increasing behavioral modes improves SS-Rate, showing that diversity broadens coverage rather than merely adding data, while removing the filter confirms that demonstrations alone cannot ensure safety. These results support the use of diverse demonstrations and test-time filtering as complementary components for robust, safe deployment. Future work will explore more effective strategies for generating diverse demonstrations and extend to broader robotic tasks.

\section*{Acknowledgements}
The authors gratefully acknowledge financial support in part by the German Academic Exchange Service (DAAD) program Konrad Zuse Schools of Excellence in Artificial Intelligence, sponsored by the Federal Ministry of Education and Research; in part by the European Union’s Horizon Europe research and innovation programme under Grant Agreement 101310328, “EXCEED-AIR”; in part by the German Federal Ministry of Research, Technology and Space (BMFTR) under the Robotics Institute Germany (RIG) funded by BMFTR grant 16ME0997K; in part by the German Research Foundation (DFG) within the RTG project ConVeY funded by grant GRK
2428; and in part by the European Union’s Horizon Europe innovation action program under Grant Agreement 101093822, “SeaClear2.0.”.



\bibliographystyle{IEEEtran}
\bibliography{reference}

\end{document}